\documentclass[letterpaper, 10 pt, conference]{ieeeconf}

\IEEEoverridecommandlockouts

\usepackage{amsmath,amsfonts}
\usepackage{algorithmicx}
\usepackage{array}

\usepackage{textcomp}
\usepackage{stfloats}
\usepackage{url}
\usepackage{verbatim}
\usepackage{graphicx}

\usepackage{balance}

\usepackage{amssymb}

\usepackage{algorithm}
\usepackage{algpseudocode}
\usepackage{siunitx}
\usepackage{array,multirow}
\usepackage{float}
\usepackage{calc}

\usepackage{tabularx}
\usepackage{hyperref}       
\usepackage{url}            
\usepackage{booktabs}
\usepackage{nicematrix,tikz}
\usepackage[LGR,T1]{fontenc}
\usepackage[greek,english]{babel}
\usepackage{xcolor}

\usepackage{threeparttable}
\usepackage{array,supertabular,hhline}
\usepackage{cleveref}

\makeatletter
\newcommand\arraybslash{\let\\\@arraycr}
\makeatother
\title{\LARGE \bf

Discovering Self-Protective Falling Policy for Humanoid Robot via Deep Reinforcement Learning
}

\author{Diyuan Shi$^{1,2}$, Shangke Lyu$^{3}$ and Donglin Wang$^{2\dagger}$ \\ {\small $^{1}$Zhejiang University \, $^{2}$Westlake University \,  $^{3}$Nanjing University \,  $^{\dagger}$Corresponding Authors }}

\begin{document}

\maketitle
\thispagestyle{empty}
\pagestyle{empty}

\begin{abstract}
Humanoid robots have received significant research interests and advancements in recent years.  Despite many successes, due to their morphology, dynamics and limitation of control policy, humanoid robots are prone to fall as compared to other embodiments like quadruped or wheeled robots. And its large weight, tall Center of Mass, high Degree-of-Freedom would cause serious hardware damages when falling uncontrolled, to both itself and surrounding objects. Existing researches in this field mostly focus on using control based methods that struggle to cater diverse falling scenarios and may introduce unsuitable human prior. On the other hand, large-scale Deep Reinforcement Learning and Curriculum Learning could be employed to incentivize humanoid agent discovering falling protection policy that fits its own nature and property. In this work, with carefully designed reward functions and domain diversification curriculum, we successfully train humanoid agent to explore falling protection behaviors and discover that by forming a `triangle' structure, the falling damages could be significantly reduced with its rigid-material body.  With  comprehensive metrics and  experiments, we quantify its performance with comparison to other methods, visualize its falling behaviors and successfully transfer it to real world platform.

\end{abstract}

\section{INTRODUCTION}
Humanoid robots have undergone tremendous developments in recent years. A diverse set of neural-network based skills has been successfully demonstrated on humanoid robots in real world with robustness and generalability.  
Remarkable examples include smooth locomotion \cite{Chen2024-sm,Zhang2025-gf}, standing from various poses \cite{Huang2025-xp,Chen2025-vb}, loco-manipulation with forceful interactions on hands \cite{Zhang2025-nr}, real-time whole-body control from human teleoperation \cite{He2024-fg,Ze2025-ij,Zhang2025-lp} and language guided whole-body motion \cite{Shao2025-lb}. These works have greatly pushed the boundary of robotic research and accelerated the adoption  of humanoid robots  in real world.

Behind the successes of these impressive works are the significant advancements of learning based methods and techniques:  Deep Reinforcement Learning (DRL) has experienced tremendous developments in recent years.  From early works like DQN \cite{Mnih2015-yw}, DDQN \cite{van-Hasselt2015-vh}   and Rainbow \cite{Hessel2017-rt}  to later methods like PPO \cite{Schulman2017-th} and SAC \cite{Haarnoja2018-ub}, the performance of DRL methods and their applications in continuous control regime have been greatly improved. Combined with tailored training framework like RMA\cite{67_kumar2021rma} and ROA \cite{Fu2022-gs} which employ teacher-student framework to better tackle Partial Observation problems  during real-world deployment, the robotic skills trained in simulation could now be transferred to real world with unprecedented easiness.

\begin{figure}[htbp]
\centering
\includegraphics[width=0.98\linewidth]{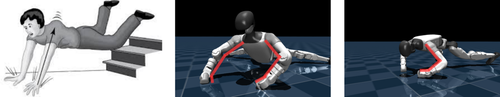}
\caption{An illustrative diagram of falling protection in human being and humanoid robot. The leftmost one depicts human's behavior and the right 2 figures are behaviors of our humanoid robot falling backward and forward, respectively. The red curves emphasize the `triangle' structure learned by our robot.}
\label{Fig_motivate}
\vspace{-10pt}
\end{figure}
However, existing works for humanoid's controlled falling are mostly based on simplified dynamics model, and either introduce strong human prior (e.g., using knees or pelvis to reduce falling impulse) or follow predefined motion trajectories \cite{Ding2018-qj, Rossini2019-xe,8206246,Li2017-pk}.  Unfortunately, humanoid robot is a \textit{high DoF system} and \textit{is mostly made of metal based rigid material} which significantly differs from human's soft body, and \textit{falling is a highly dynamic and diverse task} that could happen in various scenarios (e.g., stance or walking, various pushing directions) (more in \Cref{sect:sim_and_dissim}). Hence, these methods often struggle to cater diverse falling scenarios \cite{Rossini2019-xe,8206246} or may require additional hardware modifications to mimic human's body property \cite{Ding2018-qj,Rossini2019-xe}. As a result,  existing works have only limited evaluation in simulation, lack representative performance metrics and have limited real-world demonstrations.

To bridge this gap, in this work we firstly build a comprehensive benchmark in simulation with highly diverse falling scenarios to extensively quantify the performance of falling protection algorithms. Then we employ large scale Deep Reinforcement Learning and Curriculum Learning to incentivize humanoid robot to self-discover the most suitable falling protection policy to reduce its falling damages. Finally, we analyze the falling behaviors of our learned policy and successfully transfer it onto real-world robot platforms. In summary, our contribution could be summarized:
\begin{enumerate}
  \item We design comprehensive experiments and  representative metrics to analyze humanoid robots' falling under common scenarios like stance or walk, external perturbation and actuation failure. 
  \item With the help of Curriculum Learning (CL) and Deep Reinforcement Learning (DRL), we develop a framework to incentivize humanoid agent self-discovering falling protection policy while respecting its own physical properties.
  \item Finally, we demonstrate our policy could exhibit  sensible, self-protective behaviors (forming `triangle' structure using arms) to protect itself and be successfully transferred onto real-world robot.
\end{enumerate}

\section{Related Work}
\textbf{Learning of whole-body humanoid skills}. Humanoid robots have experienced great developments: a wide range of learned whole-body behaviors has been demonstrated on humanoid robots, including locomotion ([1], [2]), standing and recovery from diverse poses ([3], [4]), and forceful loco-manipulation [5]. Teleoperation-driven, real-time whole-body control pipelines have also been shown to generate complex behaviors on hardware ([6], [7], [8]), and recent work has begun to explore language-conditioned whole-body motion [9]. These results illustrate the growing capability and versatility of learned controllers on real humanoids.

\textbf{Deep reinforcement learning}. The evolution of DRL algorithms — from DQN, DDQN and Rainbow [10] in the early days to contemporary continuous-control methods such as PPO and SAC ([13], [14]) — has enabled more reliable policy learning for complex control tasks. Complementary training frameworks that explicitly address partial observability and domain transfer, such as the teacher–student approaches RMA and ROA ([15], [16]), have further reduced the gap between simulation and deployment, allowing simulated training to produce behaviors that transfer to real hardware with less manual tuning.

\textbf{Humanoid robot's controlled falling}. Existing approaches to humanoid falling typically adopt simplified dynamics or strong human priors, or they rely on precomputed motion trajectories ([17], [18], [19], [20]). Because humanoid robots are high-DOF, predominantly rigid systems that differ fundamentally from soft human bodies, these assumptions can limit adaptability: methods that depend on human-like compliance or limited classes of falls often fail to generalize to the broad range of real-world falling scenarios ([18], [19]), or they necessitate hardware modifications to approximate biological properties ([17], [18]). As a result, prior controlled-falling studies frequently exhibit constrained real-world evaluation.

\section{Background}
\subsection{Reinforcement Learning}
Reinforcement Learning (RL) studies the sequential decision-making problems which is formulated as a Markovian Decision Process (MDP) $\mathcal{M}=(\mathcal{S}, \mathcal{A}, \mathcal{R}, \mathcal{P}, \mathcal{S}_0, \gamma)$ where $\mathcal{S}$ is the state space, $\mathcal{A}$ is the action space, $\mathcal{R} = R(s,a) \rightarrow r$ is the reward function that outputs a scalar value given state and action, $\mathcal{P} = \mathcal{P}(s,a) \rightarrow s^\prime$ is the dynamics function that takes the current state and action then outputs the next state, $\mathcal{S}_0$ is the initial state distribution and $\gamma$ is the discount factor. In robotic system, the input space is not the full state $s$ but actually observation $o$, formulating a Partial Observational MDP. The RL's objective then is to find a policy $\pi$ that outputs action $a$ given observation $o$ to maximize total rewards:
\begin{align}
    \pi &= \pi(o) \rightarrow a = \underset{\pi}{\text{argmax}} \; \mathbb{E} \sum_{t=0}^{\infty} \bigl[\gamma^t R(s_t, a_t)\bigr].
\end{align}

In this work, we perform training in parallel simulation via Proximal Policy Optimization (PPO) \cite{Schulman2017-th} and  Lipschitz-Constrained Policies (LCP) \cite{Chen2024-sm}  with Curriculum Learning (CL) based domain randomization for enhanced Sim2Real transfer.

\subsection{Policy Learning}
In this work, we choose Unitree G1 as our testing platform. We perform Whole-Body Control (WBC) on its total 29 Degree-of-Freedom (DoF) (e.g., \textit{waists}, \textit{elbows}, \textit{knees} and \textit{wrists}). The observations are $[\dot{\omega}, r, p, q, \dot{q}, a_{t-1}]$ where $\dot{\omega} \in \mathcal{R}^3$ is the torso angular velocity, $r$ and $p$ are roll, pitch angles of the pelvis, $q \in \mathcal{R}^{29}$ and $\dot{q} \in \mathcal{R}^{29}$ are joint positions and velocity and $a_{t-1} \in \mathcal{R}^{29}$ is the action of last timestep, respectively. By following the framework of Regularized Online Adaptation (ROA) \cite{Fu2022-gs}, we feed the policy with additional privileged information during training and infer them from last 10 steps observations during deployment.

The DRL-learned policy outputs a 29-dimensional action target to the underlying PD controller which transforms it to torques and directly drives robot:
\begin{equation}
\tau(a) = K_p (a - q) - K_d \dot{q}
\end{equation}

\noindent where $\tau$ is the final torques driving the robot, $a$ is the policy output, $K_p$ and $K_d$ are the stiffness and damping coefficients of the PD controller, respectively.
The upper-level DRL policy operates in a frequency of 50Hz and the underlying PD controller operates at 200Hz.

\section{Learning to discover  Self-Protective Falling Policy}
\label{sect:method}
\subsection{Falling of human and humanoid robot}
\label{sect:sim_and_dissim}

\noindent\textbf{Similarity}. As humanoid robot mostly mimics the human being in its design, it's \textit{sensible} to mimic human's behavior when designing algorithms for  humanoid's controlled-falling. For example, when a human being is about to fall to ground accidentally, several subconscious behaviors often happen: 1) hands would be used as brace to maintain the upper-body's height to reduce a complete collision  against ground; 2) Pelvis or knee would participate to help prevent the falling; 3) Crucial body parts like head, neck and torso would be protected with maximum effort. These are all insightful observations and have inspired existing works like \cite{Ding2018-qj,Rossini2019-xe}.

\noindent\textbf{Dissimilarity}. However, falling protection of humanoid robots also have significant dissimilarity to human beings: 1) Unlike human being whose bodies are mostly covered by soft, self-healing material, humanoid robot is still made by metal based rigid material which cannot effectively absorb the falling impulse or heal after receiving damages; 2) In human's falling, the protection of different body parts naturally differ: head and neck first, torso second, pelvis, arm and legs could be sometimes `sacrificed'. But for commercial robot, the breakage on any single component (e.g., knee, pelvis or wrist) is undesirable as the robot has to stop normal functioning, wait for repair and bring financial losses. Therefore, how to \textit{maximally reduce the falling damage while respecting the humanoid robot's nature} is crucial for developing effective and suitable falling protection policy.
\begin{figure*}
\centering
\includegraphics[width=0.98\linewidth]{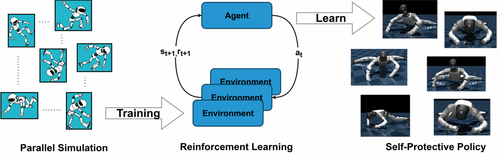}

\caption{An illustrative diagram of our learning framework.}
\label{Fig_framework}
\end{figure*}

\noindent\textbf{Our Approach}. In this work, we take a different strategy: By leveraging recent developments in parallel simulated environments and Domain Randomization, we perform large scale RL training in simulation platform (Nvidia Isaacgym \cite{Makoviychuk2021-fu})  with thousands of environment instances, each under diverse falling scenarios. Combined with carefully designed reward functions and environment diversification, we incentivize humanoid agents to explore all possible falling behaviors then choose the one that both reduces its falling damages and fits its own physical nature.  A diagram of our framework is shown in \Cref{Fig_framework}.

\noindent\textbf{Triangle Structure}. Through the training, we find our policy could successfully learn: 1) Forming a \textit{triangle structure} on  its upper arm (as in 2 right subfigures in  \Cref{Fig_motivate}), this structure could distribute falling impulse into multiple actuations and rigid bodies hence reduce falling damages while maximally protecting the arm itself from breaking; 2) Expanding the legs to further lower the CoM and reduce linear movements on the ground; 3) Protection of head, torso and pelvis to prevent critical collision on them. More analysis of its falling behaviors is shown in \Cref{sect:fall_behavior_in_simulation} and \Cref{sect:real_world}.

\begin{table*}[h]
  \centering
  \caption{Reward functions.}
  \label{tbl:reward_func}
  \begin{NiceTabular}{@{}>{\centering\arraybackslash}m{3.5cm}
                            |>{\centering\arraybackslash}m{1.75cm}
                            |>{\centering\arraybackslash}m{9cm}@{}}
    \toprule
    \textbf{Formula} & \textbf{Weight} & \textbf{Remarks} \\
    \midrule
    $\| v_{\text{root}} \|_2^2$
      & $-1.03125$
      & Penalize Root's Linear Velocity \\
    \hline
    $\max\!\big(\| c_{\text{cb}} \|_2,\,10\big)$
      & $-0.02$
      & Penalize Contact Forces on Critical Rigid bodies (Head, Torso and Pelvis) \\
    \hline
    $\| w_{\text{bodies}} \times c_{\text{bodies}} \|_2^2$
      & $-0.001$
      & Penalize All Contact Forces (contact forces on ankles are down-weighted to reduce the influence of standing contact) \\
    \hline
    $\displaystyle \mathbb{I}_{q \ge q_u}\!\big(\max(\dot{q},0)^2\big)
      + \mathbb{I}_{q \le q_l}\!\big(\min(\dot{q},0)^2\big)$
      & $-7.25$
      & Penalize Actuation Impulse \\
    \hline
    $-\min(q - q_{l},0) + \max(q-q_{u},0)$
      & $-29$
      & Penalize Out-of-Limit DoF Position \\
    \hline
    $\displaystyle \min\!\left(\frac{|\tau|}{\tau_{u}} - 0.95,\,0\right)$
      & $-24.65$
      & Penalize Out-of-Limit Driving Torque \\
    \hline
    $\tau^2$
      & $-2.9\times 10^{-3}$
      & Penalize Large Torque \\
    \hline
    $\| a_t - a_{t-1} \|_2$
      & $-\tfrac{2}{3}\times 10^{-2}$
      & Penalize Action Rate \\
    \hline
    $\| \dot{q}_t - \dot{q}_{t-1} \|_2^2$
      & $-2\times 10^{-4}$
      & Penalize DoF Acceleration \\
    \bottomrule
  \end{NiceTabular}

  \vspace{4pt} 
  \parbox{\linewidth}{\setlength{\leftskip}{6em}\footnotesize
    $\|x\|_2$ is the 2-norm. $\mathbb{I}_{\text{predicate}}(x)$ returns $x$ if the predicate is true and $0$ otherwise.
    
    $q_u$, $q_l$ are the upper and lower limits of joint positions. $\tau_u$ is the maximum driving torque.
    
    $c_{\text{bodies}}$ are contact forces of all rigid bodies. $c_{\text{cb}}$ are contact forces of critical rigid bodies.
  }
\end{table*}

\begin{table*}[h!]
  \centering
  \caption{Curriculum for training environment diversification.}
  \label{tbl:curriculum}
  \begin{NiceTabular}{@{}>{\centering\arraybackslash}m{3.5cm}
                            |>{\centering\arraybackslash}m{5cm}
                            |>{\centering\arraybackslash}m{5.75cm}@{}}
    \toprule
      \multicolumn{2}{c|}{\textbf{Setting}} & \textbf{Curriculum} \\
    \midrule

    \multirow{8}{*}{\shortstack{\; \\ \; \\ \; \\ Initial State}}
      & Initial DoF Positions
      & $\mathbb{U}(0.8q_0,\,1.2q_0)$ \\
      \hhline{~--}

    & Extra DoF Randomization
      &
        \shortstack{$q_{\text{waist\_pitch}} \sim \mathbb{U}(-0.5,0.5)$\\
                    $q_{\text{waist\_roll}}  \sim \mathbb{U}(-0.5,0.5)$\\
                    $q_{\text{waist\_yaw}}   \sim \mathbb{U}(-(p+0.2),\,p+0.2)$\\
                    $q_{\text{shoulder\_pitch}} \sim \mathbb{U}(-(p+0.4),\,p+0.4)$\\
                    $q_{\text{shoulder\_roll}}  \sim \mathbb{U}(0.15,\,1.3p+0.15)$\\
                    $q_{\text{elbow}} \sim \mathbb{U}(-0.1-0.3p,\,0.1+0.6p)$} \\
      \hhline{~--}

    & Initial Root Linear Velocity
      & $v_x,\,v_y \sim \mathbb{U}(-0.1-0.25p,\,0.1+0.25p)$ \\
    \hline

    \multirow{3}{*}{External Push}
      & Magnitude of External Push
      & $\mathbb{U}(50 + 100p,\,350 + 500p)$ \\
      \hhline{~--}

    & Direction of External Push
      & $\mathbb{U}(-\pi,\,\pi)$ \\
      \hhline{~--}

    & Body Part of External Push
      & $\mathbb{U}(\{\text{Head, Torso, Pelvis}\})$ \\
    \hline

    \multicolumn{2}{c|}{Actuation Failure}
      & \shortstack{$\mathbb{U}\bigl(\{\text{hip yaw, hip roll, hip pitch}$ \\ $\text{knee, ankle roll, ankle pitch}\}\bigr)$} \\
    \bottomrule
  \end{NiceTabular}

  \vspace{4pt}
  \parbox{\linewidth}{\setlength{\leftskip}{6em}\footnotesize
    $p$ is the training progress, which starts at $0$ and linearly increases to $1$.
    $q_0$ is the default initial DoF position.

    Domain randomizations that are not directly related to tasks are omitted for brevity.
  }

\end{table*}

\subsection{Reward Functions}
To facilitate robot exploring self-protective behavior during falling, we've carefully selected reward functions (in \Cref{tbl:reward_func}) to maintain a balance between the following objectives: 
\begin{enumerate}
  \item Reducing the damages received by robot's all body parts, joints and actuations. This is the primary goal of \textit{falling damage mitigation}.

  \item Maintain minimum human priors: Protecting head and torso where fragile and expensive devices reside.
\end{enumerate}

\begin{figure}[htbp]
\centering
\includegraphics[width=.22\textwidth]{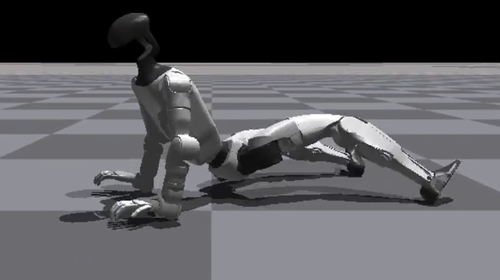}
\includegraphics[width=.22\textwidth]{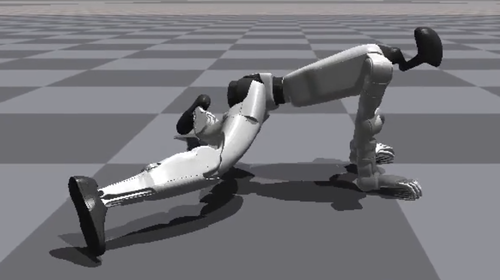}
\caption{Illustration of some behaviors unlikely to be transfered into real world deployment.}
  \label{fig:failed_cases}
\end{figure}

\noindent\textbf{Counterexample}. As we've also tried other reward functions in our preliminary experiments, most of them often result in reward-hacking behaviors: For example, we tried to directly incentivize the robot to brace using its arms, but this results in robot pushing its arms straightly against ground and is likely breaking these joints in real-world deployment as shown in \Cref{fig:failed_cases}.

\subsection{Training Curriculum}
To learn a policy that could adapt to various falling scenarios, a comprehensive curriculum of training environment diversification is also crucial. In this work, we design domain randomization curriculum from multiple angles. Specifically, external push's magnitude, direction, applied rigid body part, initial state of the robot's DoF, root velocity and the actuations to be broken are all randomized with increasing difficulty. The curriculum is listed in \Cref{tbl:curriculum}.

\section{Experiments}

\subsection{Setup}
\textbf{Environments}. Previous works in controlled falling of humanoid robots often report experimental results from only several policy rollouts which may not accurately reflect the quantitative performance of the test policy. To fill the gap, in this work we firstly build comprehensive metrics and benchmarks for assessing the falling damages of humanoid robot in various   angles and situations. An illustrative image is shown in \Cref{fig:envs}:

\begin{figure}[htbp]
\centering
\includegraphics[width=.98\linewidth]{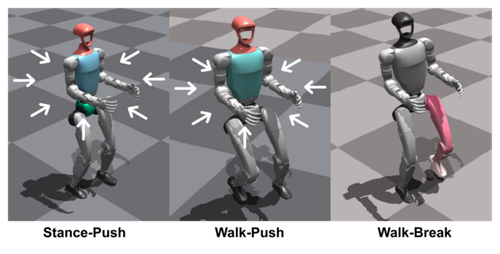}
    \caption{Illustration of our  testing environments. The white arrow represents pushing direction. The colorful body represents pushed body or failed actuation.}
      \label{fig:envs}
\end{figure}
\newlength{\strutheight}
\settoheight{\strutheight}{\strut}
\begin{enumerate}
    \item \texttt{stance-push}: The robot  stands in the beginning, then an external push is applied on it. The force magnitudes are $\{350, 550, 750\}$N, force directions are $\{0, \frac{\pi}{4}, \frac{\pi}{2}, \frac{3\pi}{4}, \pi, \frac{5\pi}{4}, \frac{3\pi}{2}, \frac{7\pi}{4}  \}$ and applied rigid bodies are $\{\text{Head}, \text{Torso}, \text{Pelvis}\}$, resulting in $3 \times 8 \times 3 = 72$ testing configurations.

  \item \texttt{walk-push}: The robot is forward walking in the beginning of the episode, then an external push is applied on it. The force magnitude is $750$N, the  force directions are $\{0, \frac{\pi}{4}, \frac{\pi}{2}, \frac{3\pi}{4}, \pi, \frac{5\pi}{4}, \frac{3\pi}{2}, \frac{7\pi}{4}  \}$, applied rigid bodies are $\{\text{Head}, \text{Torso}, \text{Pelvis}\}$ and the walking speeds are $\{0.15, 0.4, 0.75\} $m/s, resulting in $8 \times 3 \times 3 = 72$ testing configurations.

  \item \texttt{walk-break}: The robot is forward walking in the beginning of the episode, then one of its actuations is broken (the output torque of it is always zero). The broken actuations are left-$\{\text{ankle roll}, \text{ankle pitch},$ $\text{knee}, \text{hip roll}, \text{hip pitch}, \text{hip yaw}\}$, the walking speeds are $\{0.15, 0.4, 0.75\} $m/s, resulting in $6 \times 3 = 18$ testing configurations.
\end{enumerate}

\noindent \textbf{Algorithm}. We test the following algorithms in this work:
\begin{enumerate}
  \item \texttt{Baseline}: This is the original \texttt{stance} and \texttt{walk} policy trained in LCP. Despite some basic rewards to resist against external push, they are not specially trained  for falling scenario and is used as a baseline.
  \item \texttt{Zero Torque Control (ZTC)}: This is a non-controlled baseline where the robot applies zero torques during the whole falling process. This is to test the falling behavior of robot without any control.
  \item \texttt{Default Position Control (DPC)}: The policy output is always the initial pose, which resists against external perturbation via the underlying PD controller.
  \item \texttt{Ours}: This is our policy introduced in \Cref{sect:method}. It has been trained with carefully designed reward functions and curriculum to learn self-protective falling policy.
\end{enumerate}

\noindent \textbf{Metrics}. To comprehensively evaluate the falling damages of a humanoid robot, we choose the following metrics that cover energy, force and actuation aspects:
\begin{enumerate}
  \item \textit{Contact Force}: The force exerted on a rigid body when it collides against ground or other rigid bodies. This is the direct metric to represent the hardware damage that a rigid body receives during falling. 
  \item \textit{Motion Energy}: The kinetic energy of a rigid body. This is to represent the potential damages a rigid body may encounter if it collide with other objects. The value is computed as $E=\frac{1}{2} m v^2$ where $m$ is the mass of given rigid body and $v$ is the linear velocity.
  \item \textit{Actuation Impulse}: The collision between an actuation against its hardware range limiter. This could represent the damages of actuations  of the robot during falling. The value is computed as $\Bigl[\mathbb{I}(q - q_{\text{min}} \ge 0.95 \, q_{\text{range}} ) \times \text{max}(\dot{q}, 0) + \mathbb{I}(q - q_{\text{min}} \le 0.05 \, q_{\text{range}}) \times \text{min}(\dot{q}, 0)\Bigr]$ where $\mathbb{I}$ is the indicator function, $q$ is the current joint position, $q_{\text{min}}$ is the lowest possible joint position, $q_{\text{range}} = q_{\text{max}} - q_{\text{min}}$ is the range that a joint could move and $\dot{q}$ is the joint velocity.
\end{enumerate}

We further take robust statistic measures to report the final metrics:  For a given configuration, we run \textit{dozens of} repeated tests for \textit{each algorithm}, resulting in data of shape $(N, t, i)$ where $N$ is the number of repeated rollouts (98, 98, 112 for \textit{stance-push}, \textit{walk-push} and \textit{walk-break}, respectively), $t$ is the number of timesteps in each rollout (collected in 200Hz) and $i$ is the $i$th object (rigid body or joint), we take \textit{upper mean} (the average of topmost 5\

Due to the space limitation, some plots of our experimental results are presented  after \Cref{sect:conclusion}.

\subsection{Falling Damage Mitigation}
In this section we focus on the falling damages of the robot during the falling process and test if our trained policy could help mitigate these damages. 

\begin{figure}[h!]
\centering
\includegraphics[width=0.98\linewidth]{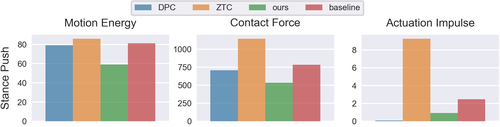}
\includegraphics[width=0.98\linewidth]{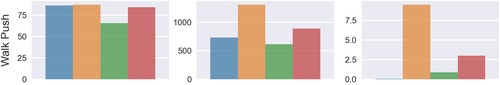}
\includegraphics[width=0.98\linewidth]{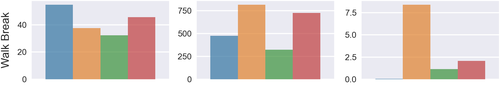}
\caption{The aggregated falling damages under various falling scenarios.}
\label{fig:fall_damage_all_metrics}
\end{figure}

\begin{figure}[htbp]
\centering

\includegraphics[width=0.98\linewidth]{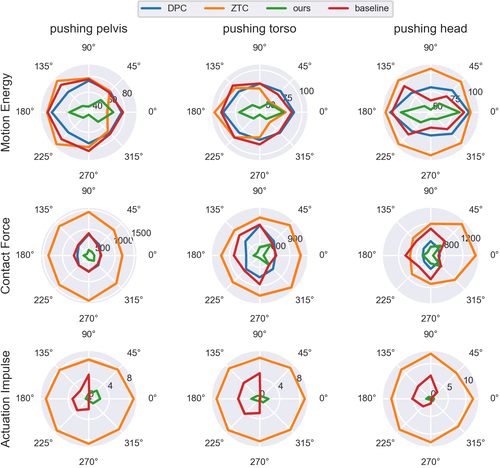}\\
\vspace{8pt}

\includegraphics[width=0.98\linewidth]{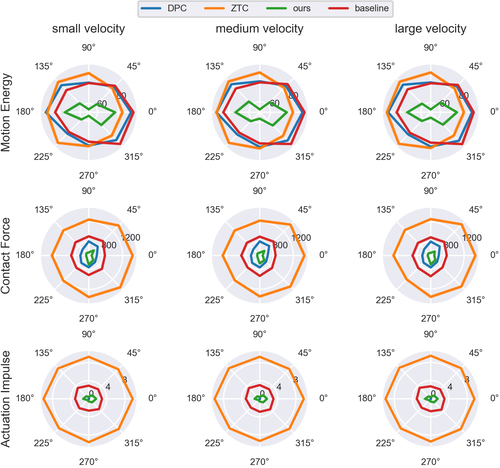}
\vspace{8pt}

\includegraphics[width=0.98\linewidth]{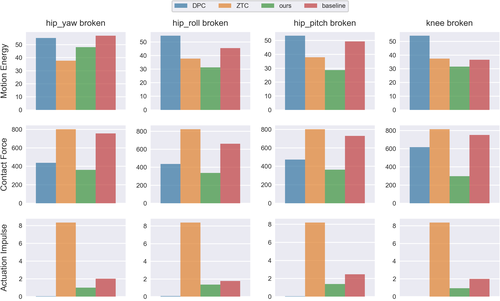}
\caption{The upper subfigures (radian plots in the first 3 rows) are falling damages of \texttt{walk-stance} on pushed rigid body. The middle subfigures (radian plots in the middle 3 rows) are falling damages of \texttt{walk-push} on walking speed. The bottom subfigures (barplots in the last 3 rows) are falling damages of \texttt{walk-break} on broken actuations.}
\label{fig:fall_damage}
\end{figure}

We show the aggregated result (averaged over all investigated factors like force magnitude, rigid body and moving speed) in \Cref{fig:fall_damage_all_metrics} and representative break-down results in \Cref{fig:fall_damage}.

From the result, we could observe that: 1) \texttt{ZTC} has the worst performance on all these 3 metrics, reflecting a completely uncontrolled falling causes severe damages to the robot; 2) \texttt{DPC} has the least \textit{Actuation Impulse} because it always outputs default pose and passively relies on the underlying PD controller's ability to resist against perturbation. On \textit{Contact Force} and \textit{Motion Energy}, it  obtains slightly better performance than \texttt{ZTC} and \texttt{Baseline}. 3) \texttt{Baseline}'s performance on \textit{Contact Force} and \textit{Actuation Impulse} is even worse than \texttt{DPC}. This (compared to the performance of \texttt{DPC}) indicates a not specially trained policy could cause even higher damages to the robot than a simple \textit{passive control}; 4) \texttt{Ours} has the best performance in \textit{Motion Energy} and \textit{Contact Force} and its slightly higher \textit{Actuation Impulse} implies an active use of certain joints to resists against falling.

\subsection{Falling Behavior Analysis}
\label{sect:fall_behavior_in_simulation}
\begin{figure}[h!]
\centering

\includegraphics[width=0.155\linewidth]{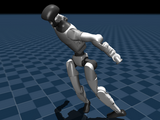}
\includegraphics[width=0.155\linewidth]{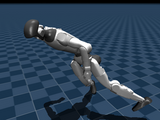}
\includegraphics[width=0.155\linewidth]{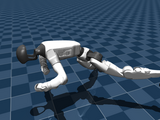}
\includegraphics[width=0.155\linewidth]{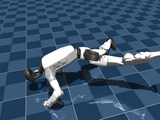}
\includegraphics[width=0.155\linewidth]{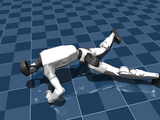}
\includegraphics[width=0.155\linewidth]{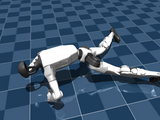}\\

\includegraphics[width=0.155\linewidth]{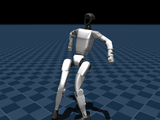}
\includegraphics[width=0.155\linewidth]{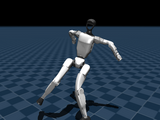}
\includegraphics[width=0.155\linewidth]{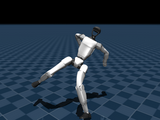}
\includegraphics[width=0.155\linewidth]{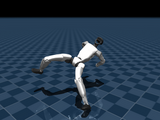}
\includegraphics[width=0.155\linewidth]{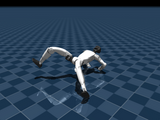}
\includegraphics[width=0.155\linewidth]{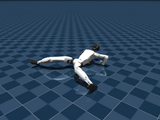}\\

\includegraphics[width=0.155\linewidth]{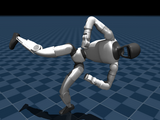}
\includegraphics[width=0.155\linewidth]{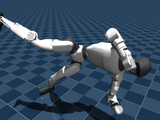}
\includegraphics[width=0.155\linewidth]{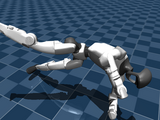}
\includegraphics[width=0.155\linewidth]{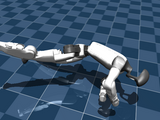}
\includegraphics[width=0.155\linewidth]{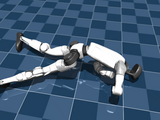}
\includegraphics[width=0.155\linewidth]{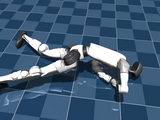}\\

\includegraphics[width=0.155\linewidth]{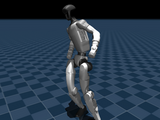}
\includegraphics[width=0.155\linewidth]{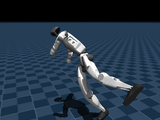}
\includegraphics[width=0.155\linewidth]{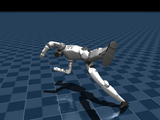}
\includegraphics[width=0.155\linewidth]{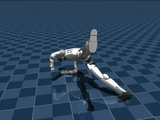}
\includegraphics[width=0.155\linewidth]{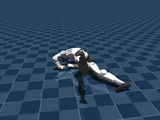}
\includegraphics[width=0.15\linewidth]{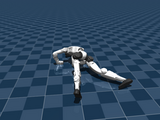}

\caption{The visualization of robot's falling behaviors. From top to bottom: front fall, back fall, left fall and right fall.}
\label{fig:fall_timestep}
\end{figure}

\begin{figure}[tbp]
\centering

\includegraphics[width=0.98\linewidth]{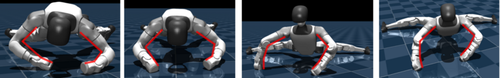}

\caption{The `triangle' structure discovered by our policy.}
\label{fig:fall_behavior}
\end{figure}

In this section, we visualize and analyze the falling behaviors of \texttt{ours} method in details. We firstly transfer our policy into Mujoco \cite{Todorov2012-bv} simulator for its better physics accuracy and take snapshoots of its falling process as images in \Cref{fig:fall_timestep} and \Cref{fig:fall_behavior}. For the video of falling in Mujoco, please refer to supplemental materials. 

In above snapshoots, it's clearly observed that our robot would actively use its arms to form an `triangle' structure to resist against falling impulse and brace the critical body parts from severe damages. Also, the legs are deliberately expanded to further lower the CoM and resist against linear movements or rolling on ground plane.

\begin{figure}[h!]
\centering
\includegraphics[width=0.98\linewidth]{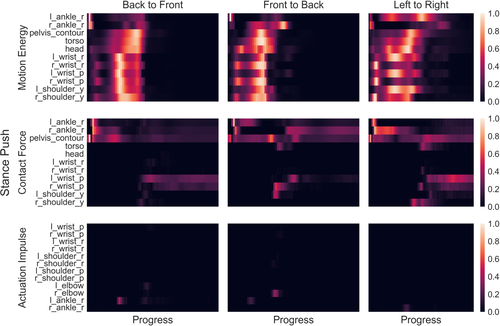} \\
\vspace{6pt}

\includegraphics[width=0.95\linewidth]{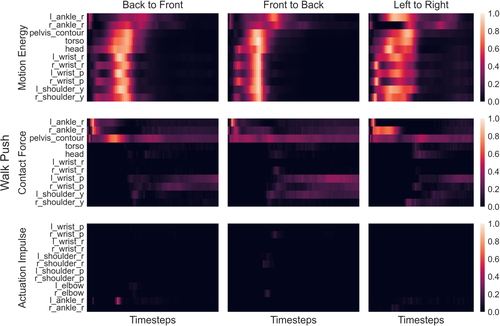}\\
\vspace{6pt}

\includegraphics[width=0.98\linewidth]{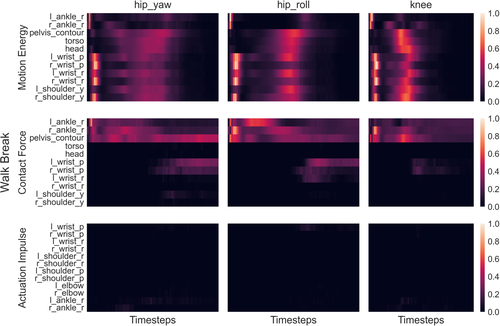}\\
\vspace{6pt}

\caption{The heatmap of falling damages of various rigid bodies or joints against timesteps of \texttt{stance-push}, \texttt{walk-push} and \texttt{walk-break}, respectively. The row in each heatmap is averaged over all falling rollouts. The prefix `l-' and `r-' represents `left' and `right' while the suffix `-r', `-p', `-y' represents `roll', `pitch' and `yaw'.}
\label{fig:fall_heatmap}
\end{figure}

Then, we plot the heatmap to further investigate the falling damages on each representative rigid body or joint in \Cref{fig:fall_heatmap}.

In \Cref{fig:fall_heatmap}, the heatmap shows the falling damages of various important rigid bodies or actuations against the timesteps where the light region represents large value while dark region represents small values. Some observations could be drawn: 1) In \textit{Motion Energy}, most rigid bodies have a peak value in first 1/3 of the process (which is the \textit{collision time}), then lose the energy due to collision. In \textit{Contact force}, the \textit{ankle roll} has contact force in the beginning of the process as the robot is standing on the ground, then loses the force due to falling. 2) The \textit{pelvis contour} receives consistent collision in all three falling modes.  As our policy is already trained to minimize the contacts on the \textit{pelvis}, this indicates that during the falling, robot's \textit{pelvis}  has large possibility to collide against ground and shall be protected with more effort. Last but not least, critical arm bodies like \textit{wrist pitch} and \textit{shoulder yaw} experience medium-level contact force after the collision time, indicating active usage of them as braces. 3) Finally, in \textit{Actuation Impulse}, most upper arm actuations exhibit neglectable impulse, despite their apparent use as braces. This indicates that `triangle' structure doesn't bring severe damages on the actuations.

\subsection{Damage Distribution Analysis}
In this section, we take a closer look at the damage distribution difference between \texttt{DPC} and \texttt{Ours} to show how these 2 methods differ in mitigating falling damage. Recall \texttt{Ours} would actively use upper arm bodies to form `triangle' structure as braces, while \texttt{DPC}  only tries to resist passively and uniformly.

\begin{figure}[htbp]
\centering
\includegraphics[width=0.98\linewidth]{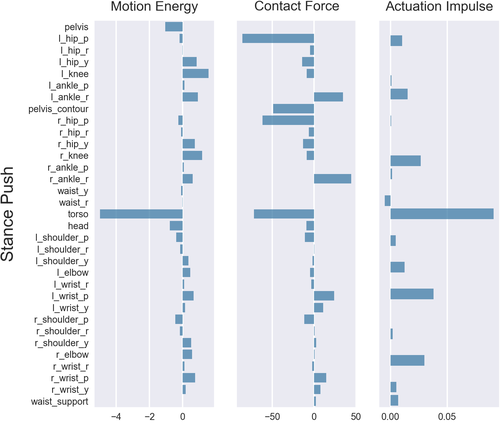}
\includegraphics[width=0.98\linewidth]{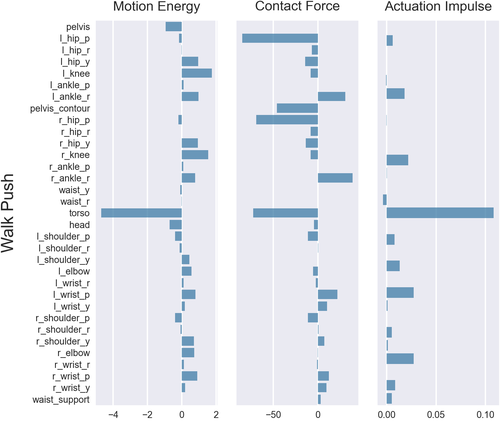}
\includegraphics[width=0.98\linewidth]{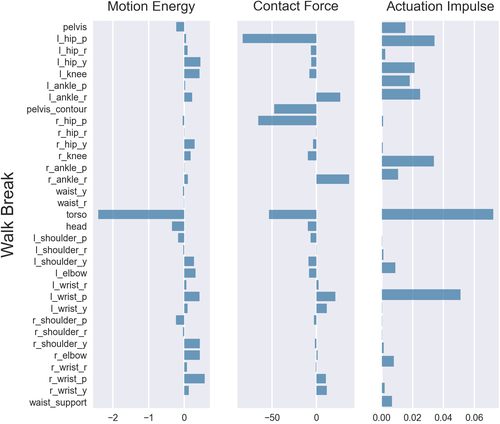}

\caption{The difference of falling damage distribution under various falling scenarios. The prefix `l-' and `r-' represents `left' and `right' respectively while the suffix `-r', `-p', `-y' represents `roll', `pitch' and `yaw' respectively.}
\label{fig:fall_damage_dist}
\end{figure}

Specifically, we plot the \textit{distribution difference} between \texttt{DPC} and \texttt{Ours} on \textit{Contact Force}, \textit{Motion Energy} and \textit{Actuation Impulse} in \Cref{fig:fall_damage_dist}. The distribution difference is computed as the absolute difference between the two categorical distributions, i.e., $D_{\texttt{ours}} - D_{\texttt{DPC}}$ where $D$ is the categorical distribution and the minus happens element-wise.

From the result, we could find: 1) In \textit{Motion Energy}, \texttt{Ours} could significantly reduce the energy on \textit{torso}, \textit{head} and \textit{pelvis} while increasing the energy on \textit{shoulder} and \textit{elbow}. This indicates that \texttt{ours} actively uses its upper arms to brace the main body; 2) Also, in \textit{Contact Force}, the contact force on \textit{head}, \textit{torso} and \textit{pelvis} are significantly reduced while \textit{wrist}s experience contact increase. This is consistent with our expectation that the robot should protect the most important body parts via the `triangle' structure; 3) In \textit{Actuation Impulse}, slight increases are observed on \textit{elbow} and \textit{wrist}s, indicating these actuations are used to form the braces with minimum hurt.  For \textit{waist pitch}, it experiences a larger actuation impulse to maintain the height and angular position of upper-body. Fortunately, the hardware  of \textit{waist pitch} is  robust enough to withstand this impulse.

\subsection{Real-world Deployment}
\label{sect:real_world}
Finally, we successfully transfer our trained policy to real-world platforms on an Unitree G1 robot. During the real world experiment, we run a pre-trained walking policy on the robot and make it walk in place (This differs from standing still as the legs of our robot are still moving). Then a human researcher deliberately pushes the robot on the torso to make it fall. At the same time, we switch to the falling policy which runs to reduce the falling damage of robot. A series of snapshoot images are in \Cref{fig:real_world_fall} and the  videos are in supplemental materials.

\begin{figure}[h!]
\centering
\includegraphics[width=0.18\linewidth]{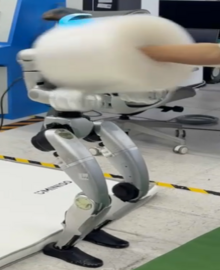}
\includegraphics[width=0.18\linewidth]{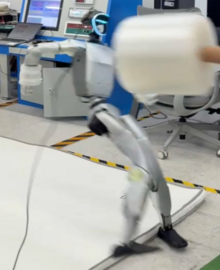}
\includegraphics[width=0.18\linewidth]{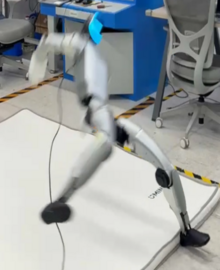}
\includegraphics[width=0.18\linewidth]{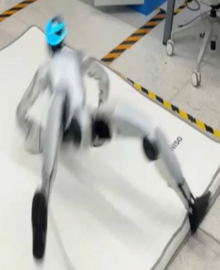}
\includegraphics[width=0.18\linewidth]{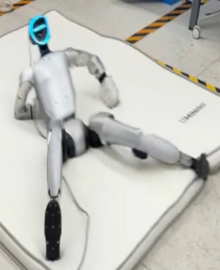} \\  

\includegraphics[width=0.18\linewidth]{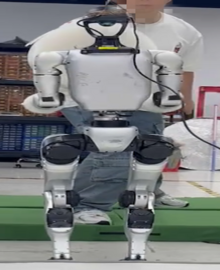}
\includegraphics[width=0.18\linewidth]{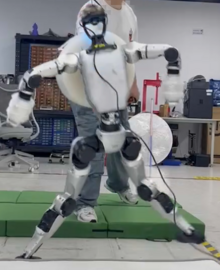}
\includegraphics[width=0.18\linewidth]{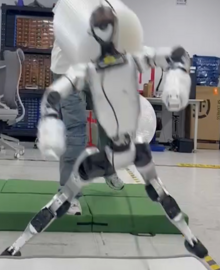}
\includegraphics[width=0.18\linewidth]{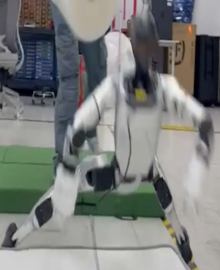}
\includegraphics[width=0.18\linewidth]{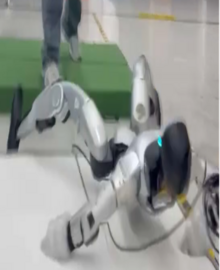}

\caption{The image snapshoots from our real-world deployment. The robot is pushed to fall and it successfully braces itself with `triangle' structure. Please view videos in our supplemental materials for better visualization.}
\label{fig:real_world_fall}
\end{figure}

From the images and videos, it's clear that our robot is able to brace itself when pushed to fall against ground and protect its critical bodies with `triangle' structure similar in simulation. Also, this process happens when the robot is walking which is a dynamic and diverse situation.

\section{Conclusion}
\label{sect:conclusion}

Despite the growing research interests and technical advancements in humanoid robots, their inclination to falling has greatly hindered their useage in both academia and industry.  In this paper, we study this issue from 2 angles: Firstly we design various  metrics and experiments to thoroughly analyze the falling damages of humanoid robots; Then we employ DRL and CL techniques to train a  policy which successfully mitigates the falling damages by discovering a `triangle' structure that fits its body dynamics best. 

In the future work, we hope to further explore the effectiveness of our method in more diverse falling scenarios and augment our method with perception to the surrounding environments such that it could fall with proper interactions with nearby objects.

\newpage

\rule{0pt}{98pt}

\balance

\bibliographystyle{IEEEtran}
\bibliography{IEEEexample}

\end{document}